\documentclass[letterpaper, 10 pt, conference]{ieeeconf}

\IEEEoverridecommandlockouts                              

\usepackage{hyperref}
\usepackage{amsmath}
\usepackage{subcaption}
\usepackage{graphicx}
\usepackage{verbatim}
\usepackage{amsfonts}
\usepackage[noend]{algpseudocode}
\usepackage{algorithm}
\usepackage{multicol}
\usepackage{bm}
\usepackage{pifont}
\usepackage{cite}

\title{\LARGE \bf
Real-time Quadrotor Navigation Through Planning in Depth Space in Unstructured Environments
}

\author{Shakeeb Ahmad$^{1}$ and Rafael Fierro$^{1}$
\thanks{This work was supported in part by Air Force Research Laboratory (AFRL) under agreement number FA9453-18-2-0022.}
\thanks{$^{1}$Department
of Electrical and Computer Engineering, University of New Mexico, Albuquerque,
NM, 87106 USA 
        {\tt\small \{shakeebahmad,rfierro\}@unm.edu}}
}

\begin{document}

\maketitle
\thispagestyle{empty}
\pagestyle{empty}

\begin{abstract}

This paper addresses the problem of real-time vision-based autonomous obstacle avoidance in unstructured environments for quadrotor UAVs. We assume that our UAV is equipped with a forward facing stereo camera as the only sensor to perceive the world around it. Moreover, all the computations are performed onboard. Feasible trajectory generation in this kind of problems requires rapid collision checks along with efficient planning algorithms. We propose a trajectory generation approach in the depth image space, which refers to the environment information as depicted by the depth images. In order to predict the collision in a look ahead robot trajectory, we create depth images from the sequence of robot poses along the path. We compare these images with the depth images of the actual world sensed through the forward facing stereo camera. We aim at generating fuel optimal trajectories inside the depth image space. In case of a predicted collision, a switching strategy is used to aggressively deviate the quadrotor away from the obstacle. For this purpose we use two closed loop motion primitives based on Linear Quadratic Regulator (LQR) objective functions. The proposed approach is validated through simulation and hardware experiments.

\end{abstract}
\section{INTRODUCTION}
Quadrotor aerial navigation using vision is gaining a lot of importance recently. One of the motivations behind this paper is to mimic the `Race the Sun' game \cite{racethesun}. The game involves a spaceship navigating forward and consuming solar energy while the sun slowly sets over the horizon. Moreover, it can only exploit limited maneuvers while flying owing to the vehicle dynamics and speed. Such a set up has important implications on practical systems. In this kind of scenario, the aircraft has to undergo maneuvers that are fuel efficient. Moreover, it has to plan its trajectories using a limited reachability set as a result of limited field of view of the onboard sensors such as a camera. Recently, Lockheed Martin partnered with the ESPN's Drone Racing League for the AlphaPilot Innovation Challenge \cite{droneracingleaguenews}. The objective is to develop an artificial system that is able to compete with the human pilots while navigating through a three-dimensional obstacle course. Accomplishing this task in real-time primarily requires efficient processors on the hardware side and fast decision-making algorithms on the software side. 
\subsection{Related Works}
Some earlier works on real-time trajectory generation by solving optimization and performing cost function update onboard include \cite{scherer2009efficient} and \cite{ratliff2009chomp}. As for the vision-based robotic navigation, mapping based approaches are one of the most commonly used methods. Simultaneous localization and mapping (SLAM) techniques helped achieve real-time perception for efficient decision making \cite{huh2013integration, alzugaray2017short-term, bachrach2011range}. Authors in \cite{cigla2017image} used a more efficient technique relying on temporary short-term maps known as ego-centric cylinders, rather than saving and creating full map of the environment. Similarly, \cite{schmid2013stereo} proposed a technique to solve the problem of navigation in complex environments using onboard processing and 3D maps. This research enabled development of better algorithms for real-time navigation. 

One of the main problems of concern in robotic vision is that an RGB image from a camera provides information only for the two axes. Calculation of distances or the values associated with the third axis provides more information to the robot in vision aided navigation. Authors in \cite{lamers2016self} proposed an approach to self-learn relative distances onboard to facilitate this kind of navigation. One way to obtain depth is to use a well known stereo pair of cameras to calculate the pixels' depth in an image. This computation is governed by the epipolar geometry and triangulation. \cite{ghosh2017joint} used a stereo camera to compute depth in an event-triggered way. This approach reduces unnecessary computations of depth maps when a robot has limited hardware resources for processing. From the planning perspective, many optimization techniques might fail in real-time and/or onboard a robotic agent because of their computational complexity. A library-based approach is known to be quite effective to fast navigation. This technique allows to compute some limited motion primitives off-line prior to the flight. Some previous work in this area include \cite{mueller2015computationally} and \cite{barry2018high}. \par
The authors in \cite{matthies2014stereo} claimed to be the first ones to come up with the C-space expansion technique for the depth images. The depth image is modified and the scene information undergoes an expansion to allow the quadrotor to be treated as a point mass for collision checks. They used rapidly exploring random trees (RRT) for path planning. Later, authors in \cite{dubey2017droan} extended the approach for obstacle avoidance in disparity space. Authors of \cite{barry2018high} used a library of seven pre-computed trajectories which are appended and executed onboard. They assumed that the patches of the feasible trajectory started from the same initial condition \textit{i.e.,} a perfect hover. The stability analysis was also out of scope of their work. Other contributions to the area of vision-based navigation of UAVs include \cite{liu2017search, mohta2018fast, loianno2018special}. In \cite{mohta2018fast}, the authors used an A* algorithm for path planning with local and global map information. They also performed trajectory optimization online. Some other contributions in the area of visual perception and trajectory planning for obstacle avoidance are reported in \cite{perez2018architecture, faessler2016autonomous, allen2016real, aditya2016motion}. They dealt with the problem of aerial navigation in GPS-denied environments. Authors in \cite{faessler2016autonomous} focused on mapping based techniques while \cite{aditya2016motion} based their approach on certain pre-defined motion primitives for fast flight. A reachability-based approach for planning in the presence of noise while performing replanning using an efficient self-triggered way is explored in \cite{yel2017reachability}.  \par

\subsection{Contributions}
There has been an extensive work in the area of real-time navigation in the presence of obstacles at known locations or in the area of vision-based planning using mapping-based techniques for slower navigation. However, a complete solution for fast real-time vision-based navigation in unstructured environments still requires significant research contributions. We attempt at bridging some gaps in the research, by designing a framework for a stereo vision-based agile quadrotor navigation for real-time and onboard computations. To achieve this goal, we adopt a collision detection strategy \cite{smith2017pips} which requires minimal amount of computations without relying on building maps or processing the depth image for C-space expansion. To the best of our knowledge, this technique is still unexplored for fast navigating robots such as a UAV. Moreover, we design our trajectory generator using LQR-based close loop motion primitives. This technique does not require to search the whole 3D space unlike sampling based algorithms or reachability based graph search algorithms. Moreover, the trajectories generated using this approach are dynamically feasible and optimal with respect to predefined quadratic costs. Also our technique does not require to assume hovering initial condition to generate trajectories for each time horizon, unlike many other methods proposed in the literature for vision-based navigation. Finally, the technique is tested on an actual quadrotor UAV with all the processing and sensing done onboard the quadrotor UAV. The proposed hardware design is also discussed in this paper. Our active perception framework for UAVs can be used with a variety of other planning algorithms such as sampling based and graph search methods.  \par

\subsection{Limitations} 
Some limitations that we consider for our setup are mentioned as follows. The workspace structure is unknown \textit{i.e.,} the quadrotor is not provided with any obstacle information before flight. The quadrotor localization solely relies on the visual odometry from a stereo camera and the state estimate from the Inertial Measurement Unit (IMU). The same stereo camera is used to perform obstacle detection. This camera is rigidly attached to the quadrotor body and is facing forward with respect to the quadrotor body, at all times. The quadrotor generates the trajectories in the look-ahead fashion. To generate a look-ahead trajectory, the quadrotor has to perform collision checks. However, these collision checks can only be performed inside the camera's field of view. Since the camera is rigidly attached to the quadrotor body, the quadrotor only sees a limited area to perform collision checks, which changes as the quadrotor moves. This means that the quadrotor's field of view is a function of its pose and time and so is the reachability for the look-ahead trajectory. This is explained in detail later. Moreover, the planner has to make sure that the generated trajectories are dynamically feasible and follow the pre-defined quadratic cost functions to be fuel optimal.

\subsection{Organization}
The rest of the paper is organized as follows. Section \ref{sec:visual_perception} presents the real-time vision-based collision detection strategy. Section \ref{sec:_trajectory_generation} discusses the trajectory generation methodology which exploits optimal control and the proposed collision detection methodology. Section \ref{sec:implementation} discusses the implementation details including the simulation and hardware experiments. Finally, Section \ref{sec:conclusion} provides concluding remarks.

\section{VISUAL PERCEPTION}
\label{sec:visual_perception}
 Let $\mathcal{F_W}$ be the reference frame fixed with the workspace $\mathcal{W}  \subset \mathbb{R}^3 $. The problem setup includes obstacles $\mathcal{O}_i \subset \mathcal{W}$, of unknown geometries, locations and distribution in a workspace $ \mathcal{W}$. The workspace is assumed to be compact. Let, $\mathcal{F_B}$ be the reference frame attached to the quadrotor body. To perform the relative coordinate transformations between the frames, the knowledge of the quadrotor configuration is required. Let $ \bm{q} = (x,y,z,\phi,\theta,\psi) \in \mathcal{C} $ represent a configuration of the quadrotor in its configuration space $ \mathcal{C} $. Here $\phi,\theta$ and $\psi$ are roll, pitch and yaw angles of the robot respectively. Moreover, we let $ \mathcal{A}(\bm{q}) \subset \mathcal{W} $ and $\mathcal{G} (\bm{q}) \subset \mathcal{W} $  be expressed as a set of all the points occupied by the quadrotor and the camera's field of view respectively, while the quadrotor maintains a certain configuration $ \bm{q} $. The collision-free state is represented as $ \mathcal{C}_{free} = \mathcal{C} \backslash \mathcal{C}_{obs} $. The configuration for which the quad-rotor collides with the obstacle is given as $ \mathcal{C}_{obs} = \{ \bm{q} \in \mathcal{C} : \mathcal{A}(\bm{q}) \cap \mathcal{O}_i \neq \phi \}$.

\subsection{Collision Detection}
\label{sec:_perception_strategy}

\begin{figure}[!ht]
    \centering
    \includegraphics[width = \linewidth]{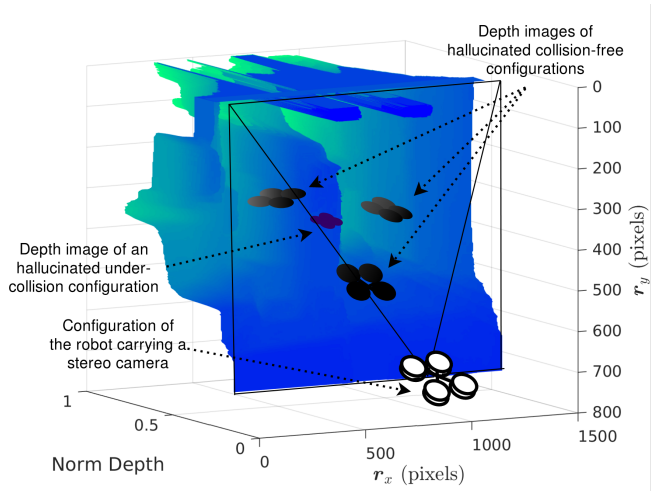}
    \caption{Comparison of the 3D scene depth image (in blue) with the depth images of the hallucinated robots (in black). }
    \label{fig:my_label}
\end{figure}

 \begin{figure*}[!ht]
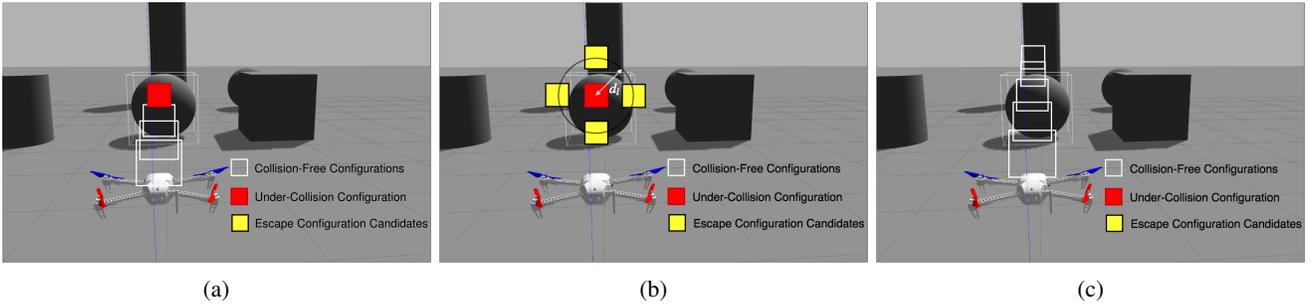

  \begin{subfigure}[b]{0.32\textwidth}
    \includegraphics[width=\textwidth]{figures/escape00.png}
    \caption{}
  \end{subfigure}
  \begin{subfigure}[b]{0.32\textwidth}
    \includegraphics[width=\textwidth]{figures/escape01.png}
    \caption{}
  \end{subfigure}
    \begin{subfigure}[b]{0.32\textwidth}
    \includegraphics[width=\textwidth]{figures/escape02.png}
    \caption{}
  \end{subfigure}
  \caption{\label{fig:escape_point_query} Escape configuration computation (Squares are used to represent the configurations for simplification). (a) Collision is detected in a look-ahead  trajectory configuration. (b) Four candidate configurations, around the under-collision configuration, are checked for collisions. (c) Collision-free trajectory is generated towards the escape configuration. }
\end{figure*}

\textbf{Problem 1:} \emph{Let the quadrotor be following a trajectory $\bm{q}$. Given a quadrotor configuration $\bm{q}(t_c) \in \mathcal{C}_{free}$ at any time instant $t_c$, find whether a configuration $\bm{q}(t)$ is under collision, collision free or outside the field of view for $t > t_c$.  }

We use the notion of planning in depth image space to solve this problem. While following a trajectory $\bm{q}$, the quadrotor is able to perform the collision checks at any time instant. In order to continue following the trajectory $\bm{q}$, the robot needs to know if the look-ahead trajectory is under collision or not. To perform this check, the quadrotor hallucinates itself along the given trajectory $\bm{q}$ in $\mathcal{F_W}$, forward in time. The collision detector creates the depth images of the hallucinated quadrotor configurations and compares it with the depth of the actual 3D scene as seen by the forward facing stereo camera. If the depth image of an hallucinated configuration exceeds the depth of the actual scene, then the configuration is under collision. If the hallucinated configuration does not project to valid image points, then it is outside the field of view of the quadrotor. 

\par Let us consider that the quadrotor is following a trajectory of configurations $\bm{q}$, and has reached a configuration $\bm{q}(t_c)$. The planner wants to know if a hallucinated configuration $\bm{q}(t > t_c)$ is under collision or not. The first step is to create a depth image of the configuration $\bm{q}(t)$. Following the conventional graphics methodology, the depth image of a 3D scene model is generated by projecting the 3D objects models to the image and selecting, for each pixel, the closest point to the camera. However, we will generate the hallucinated robot's depth image by projecting the farthest points of the robot sitting at a configuration $\bm{q}(t)$ \cite{smith2017pips}. 

We use a pinhole model for the stereo camera. Let $\bm{P}_\mathcal{W} = (x_w, y_w, z_w) \in \mathbb{R}^3$ be a 3D scene point in $\mathcal{F_W}$. Before projecting it to an image, it has to be transformed to the body frame $\mathcal{F_B}$ and then to the camera coordinate frame $\mathcal{F_S}$. Let the point $\bm{P}_\mathcal{W}$ in the camera coordinate frame be $\bm{P}_\mathcal{S} = (x_s, y_s, z_s) \in \mathbb{R}^3$. The transformation among the corresponding homogeneous coordinates is governed by the following expression.

\begin{align}
    \label{eq:transformation1}
    \begin{bmatrix}
    x_s \\ y_s \\ z_s \\ 1
    \end{bmatrix} =
    \begin{bmatrix}
    \bm{R}_\mathcal{W}^\mathcal{S}(\phi, \theta, \psi) & \bm{T}(x,y,z) \\
    \bm{0}_{3\times3} & 1
    \end{bmatrix}
    \begin{bmatrix}
    x_w \\ y_w \\ z_w \\ 1
    \end{bmatrix}.
\end{align}

The scene point in the camera coordinates is then projected to the image using the following expression.

\begin{align}
\label{eq:transformation2}
\bm{r} \sim \begin{bmatrix}
fs_x & 0 & c_x & 0 \\
0 & fs_y & c_y & 0 \\
0 & 0 & 1 & 0
\end{bmatrix}
\begin{bmatrix}
x_s \\ y_s \\ z_s \\ 1
\end{bmatrix}.
\end{align}

The above expression essentially converts the homogeneous point coordinates to the homogeneous image coordinates. Here $\bm{R}_\mathcal{W}^\mathcal{S} \in SO(3)$ and $\bm{T} \in \mathbb{R}^3$ represent the rotation and translation matrices for the transformation from $\mathcal{F_W}$ to $\mathcal{F_S}$. The rotation from $\mathcal{F_W}$ to $\mathcal{F_B}$ is given by the Z-X-Y rotation matrix $\bm{R}^\mathcal{B}_\mathcal{W} \in SO(3)$ while the rotation from $\mathcal{F_B}$ to $\mathcal{F_S}$ is governed by $\bm{R}^\mathcal{S}_\mathcal{B} = \bm{R}_y (- \pi / 2) \times \bm{R}_x (\pi / 2) \in SO(3) $. This transformation (and hence projection) clearly depends on the quadrotor configuration at time instant $t_c$. Therefore, we need to feedback the quadrotor's most recent configuration to the collision detector whenever the collision check needs to be performed. The intrinsic camera calibration parameters are the pixel dimensions $s_x$ and $s_y$, the location of the camera optical axis $c_x$ and $c_y$ and the focal length of the camera lens $f$. Finally, $\bm{r} = (r_x, r_y) \in \mathbb{R}^2$ is the pixel location of the transformed point in image coordinate frame. \par 
In an RGB camera the information at a pixel location $\bm{r}$ corresponds to the color of the 3D scene point that is projected to $\bm{r}$. These kind of images lose the information in the $z$-axis of the camera coordinate frame $\mathcal{F_S}$. However, in a depth image, the information stored at the location $\bm{r}$ is the lost $z$-axis distance information (in $\mathcal{F}_\mathcal{S}$) of the projected 3D point. Given the camera pose (and hence the quadrotor configuration at $\bm{t}_c$), we can compare the location of an hallucinated configuration $\bm{q}(t > t_c)$ with the depth of the environment. However, we do not know which pixels correspond to the hallucinated quadrotor for comparison. We create a depth image of the hallucinated robot and compare it with the depth image of the actual 3D scene from the stereo camera. 

To create a depth image from an hallucinated quadrotor, we assume that it is sitting at a configuration $\bm{q}(t)$ in an empty world. Given the mesh model of the quadrotor in an empty world, let $\{\bm{P}_\mathcal{W}^i\}, i = 0,1,2,3,..., $ be the set of all points at the surface of the quadrotor projecting to a pixel $\bm{r}^j$. After transforming these points to the camera coordinates, they are represented as $\{\bm{P}_\mathcal{S}^i\}, i = 0,1,2,3,...$. In a conventional computer vision approach, only that point wins which is closest to the camera. However, while creating the depth image of the hallucinated robot we select the point $\bm{P}_\mathcal{S}^j$ to be projected to the image at the pixel $\bm{r^j}$ where,
\begin{align}
     j = \textnormal{arg} \max _i \: z_s^i.
\end{align}

For any point $\bm{P}_\mathcal{W}^j$ (or $\bm{P}_\mathcal{S}^j$), if the pixel location $\bm{r^j}$ is not a valid image coordinate then the hallucinated quadrotor is considered outside the field of view. More formally, 
\begin{align*}
    \textnormal{if} \quad \exists \: \: j \textnormal{ such that } \bm{r}^j \notin \mathcal{R} \quad \textnormal{then} \quad \mathcal{A}(\bm{q}(t)) \not \subset \mathcal{G}(\bm{q}(t_c)).
\end{align*}

Here, $\mathcal{R} \subset \mathbb{R}^2$ is a set of valid image coordinates. Since the camera is rigidly attached to the quadrotor body, the set $\mathcal{G}(\bm{q}(t_c))$ changes with the quadrotor pose and hence with the time when the collision check is performed. \par

If inside the field of view, the quadrotor configuration is checked for collision, \textit{i.e,}
\begin{align*}
    \textnormal{if} \quad z_s^j > D_{\bm{r}^j} \: \forall \: \bm{r}^j \in \mathcal{R} \: \: \textnormal{and} \: \: \mathcal{A}(\bm{q}(t)) \subset \mathcal{G}(\bm{q}(t_c)) \nonumber \\
    \textnormal{then} \quad \bm{q}(t) \in \mathcal{C}_{free}.
\end{align*}

Here $D_{\bm{r}^j}$ is the depth of the actual 3D scene at the pixel $\bm{r}^j$. If both of the above conditions are not satisfied, the configuration $\bm{q}(t)$ is declared under collision.

 \subsection{Finding Escape}

 If any hallucinated configuration $\bm{q}(t>t_c)$ is under collision, higher depth areas are checked for collision, parallel to the image plane. The quadrotor is hallucinated in the up, down, left and right direction on a circle of radius $d_l$, around the configuration $\bm{q}(t)$. These configurations are checked for collision to see if the quadrotor can fit through the gap in the 3D scene. If a collision-free configuration is found, it is selected as an escape configuration $\bm{q}_{esc} \in \mathcal{C}_{free}$. If all four configurations are under collision, the process is repeated by checking more configurations in up, down, right and left directions at radii $2 d_l$. The process repeats for the circle radii $3 d_l, 4 d_l, ...$, until either a collision-free configuration is found or the candidate configurations in all the directions leave the field of view. In the later case, the robot is assumed to be stuck under its field of view constraints. This happens if the field of view of the quadrotor is completely blocked, leaving no gaps to escape through as shown in Figure \ref{fig:escape_point_query}. We look for escapes in a deterministic fashion. However, a more random approach can be adopted to look for the escapes by performing a denser search.

\section{OPTIMAL CONTROL}

Differential flatness is a useful property exhibited by a quadrotor UAV system \cite{mellinger2012trajectory}. Flat outputs of a differential flat system determine the behaviour of the system. This allows us to plan the trajectories in the output space which can be mapped back to the approximate inputs. Differential flatness property for the quadrotor UAVs allows us to decouple the inputs for the three position axes. We will exploit this property for the trajectory generation to save online computations. Let,
\begin{align}
    \label{eq:state_vector}
    \bm{x}(t) = [\bm{p}(t)^T, \dot{\bm{p}}(t)^T, \ddot{\bm{p}}(t)^T, ..., \quad \bm{p}^{(n-1)}(t) ] \in \mathcal{X},
\end{align} 
be the quadrotor system states, represented as 3D position and its $(n-1)$ derivatives.  Here, $\bm{p}(t) = (x,y,z) \in \mathcal{W}$ is the 3D quadrotor position at the time instant $t$. The set of valid system states is represented by $\mathcal{X} \subset \mathbb{R}^{3n}$. Let $\mathcal{U} \subset \mathbb{R}^3$ be the set of admissible inputs to the system. Choosing $n=2$ for the state space (Equation (\ref{eq:state_vector})), the system can be controlled by generating the trajectories using the following relation. 
\begin{align}
    \label{eq:doubleint_model}
    \bm{\bm{\dot{p}}(t)} = \bm{u}(t), \quad \bm{u}(t) \in \mathcal{U}.
\end{align}

This can be written in the standard state space form as,
\begin{align}
    \label{eq:doubleint_model_statespace}
    \bm{\dot{x}}(t) = \bm{A}\bm{x}(t) + \bm{B}\bm{u}(t),
\end{align}
where,
\begin{align*}
    \bm{A} = \begin{bmatrix}
    \bm{0}_{3\times3} & \bm{I}_{3\times3} \\ \bm{0}_{3\times3} & \bm{0}_{3\times3}
    \end{bmatrix}, \quad
    \bm{B} = \begin{bmatrix}
    \bm{0}_{3\times3} \\ \bm{I}_{3\times3}
    \end{bmatrix}.
\end{align*}

We use the model shown in Equations (\ref{eq:doubleint_model}) and (\ref{eq:doubleint_model_statespace}) for generating the trajectories. However, our collision detector takes as an input, the configuration of the robot that is performing the collision check as well as the configuration of the hallucinated robot that is to be checked for collision. These configurations are referred to as $\bm{q}(t_c)$ and $\bm{q}(t)$ respectively. These configurations include the position and orientation information. The onboard state estimator provides the information about the most recent configuration $\bm{q}(t_c)$ of the actual robot. Since our trajectory generator does not generate trajectories in $SE(3)$, we do not know the orientation of the hallucinated robot sitting at a position $\bm{p}(t)$ on a look-ahead trajectory. To check a robot position $\bm{p}(t)$ for collision we simply assume the most pessimistic robot orientation. Keeping in view the maximum roll and pitch angles of the quadrotor, this is the configuration such that we can project the quadrotor to the maximum number of pixels in the image. Consequently, $\bm{x}(t) \in \mathcal{X}_{free}$ if the most pessimistic configuration of the quadrotor (that is sitting at $\bm{p}(t)$) is collision free. Similarly, the escape state $\bm{x}_{esc} \in \mathcal{X}$ corresponds to the configuration $\bm{q}_{esc} \in \mathcal{C}$.

\subsection{Trajectory Generation}
\label{sec:_trajectory_generation}
\textbf{Problem 2.} \emph{Let the collision free states be represented as $\mathcal{X}_{free} \subset \mathcal{X}$.
Given an initial state $\bm{x}_0 \in \mathcal{X}_{free}$ and a goal region $\mathcal{X}_{goal} \subset 
\mathcal{X}_{free}$, find a trajectory $\bm{x} : [0\quad t_f] \to \mathcal{X}$} such that: 
\begin{align*}
    \mathop{\textnormal{minimize}}_{\bm{u,x}} \: \: &s(t)J_0(\bm{u,x}) + (1-s(t))J_1(\bm{u,x}) \\
    \textnormal{subject to} \: &\dot{\bm{x}}(t) = \bm{A}\bm{x}(t) + \bm{B}\bm{u}(t), \quad \forall \: t \in [0\quad t_f] \\
                             &\bm{x}(0)=\bm{x}_0, \quad \bm{x}(t_f) \in \mathcal{X}_{goal}  \\
                             &\bm{x}(t) \in \mathcal{X}_{free}, \quad \bm{u}(t) \in \mathcal{U}, \quad \forall \: t \in [0 \quad t_f].
\end{align*}

\begin{figure}
\centering
\includegraphics[width=0.48\textwidth]{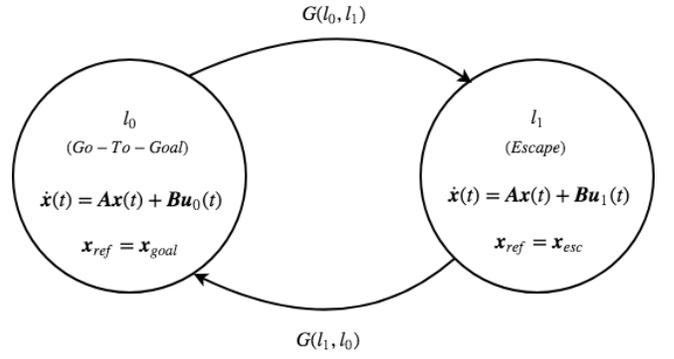}
\caption{\label{fig:hybrid_automaton} The hybrid automaton.}
\end{figure}

Here $s$ is the binary variable used to perform the switching between the two pre-defined quadratic objectives, $J_0(\bm{u,x})$ and $J_1(\bm{u,x})$. The first control objective takes the quadrotor forward towards the goal with emphasis on saving energy, while the second one is designed for quick sharp turns around the obstacles to see what is behind and past the obstacle. These control objectives lead to different closed loop dynamics at various instances of the maneuver. Each control objective corresponds to a discrete mode. Since the dynamics of the system are continuous and the modes are discrete, the system can be represented as a hybrid automaton \cite{van2000introduction} as shown in Figure \ref{fig:hybrid_automaton}. A discrete mode $l_i$ corresponds to the $i$th control objective while a guard $G(l_i, l_j)$ defines a state for which the switching has to occur from mode $l_i$ to $l_j$. 

\begin{figure}
    \centering
    \includegraphics[width = \linewidth]{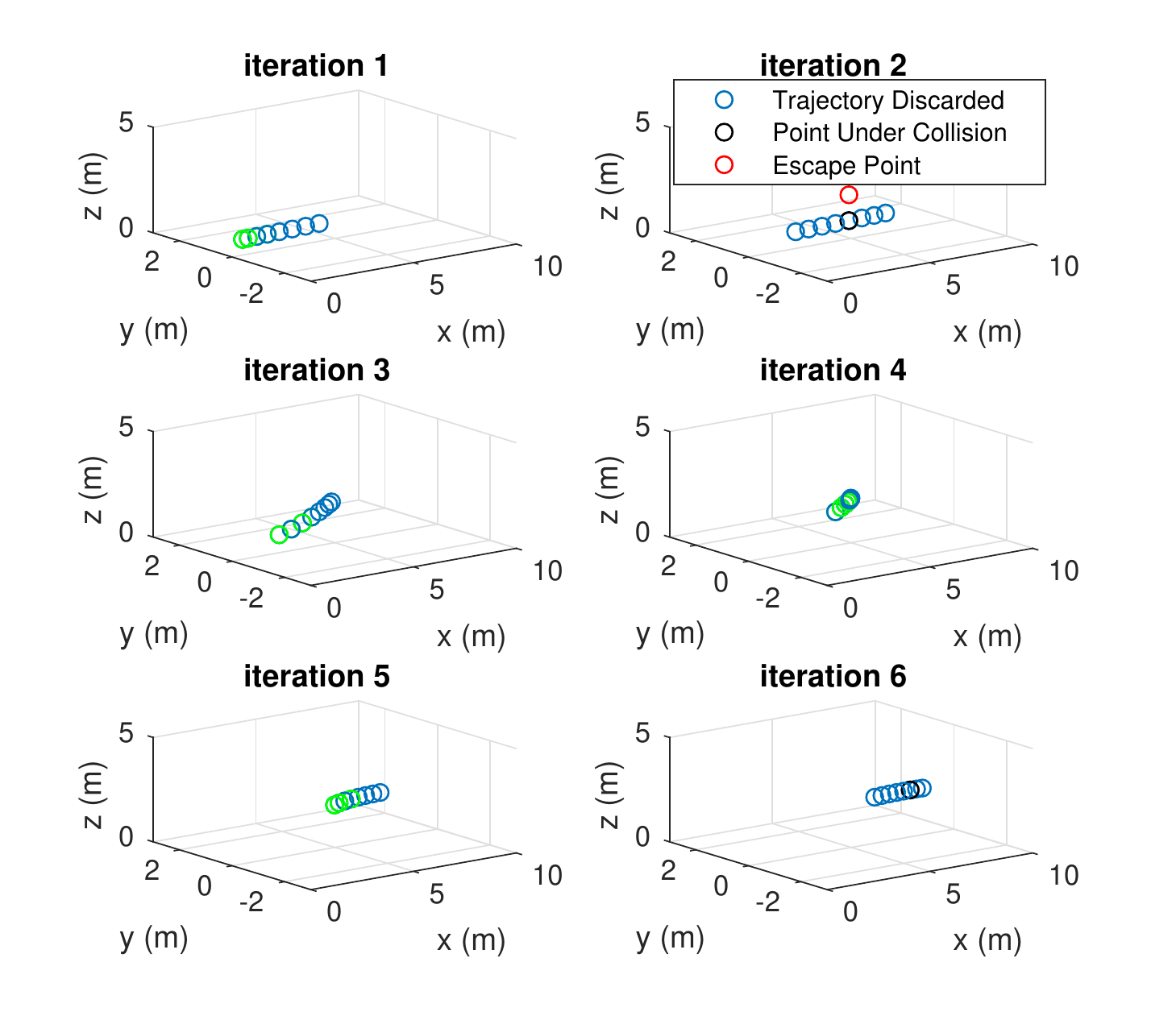}
    \caption{Cartoon to show the trajectory generation method.}
    \label{fig:traj_gen_cartoon}
\end{figure}

The workspace is assumed to be unknown and unstructured. Moreover, it can only be explored by the stereo camera attached to the quadrotor body. Therefore, the final trajectory is obtained by appending the small fixed horizon look-ahead trajectories. Let the $k$th look-ahead trajectory be $\bm{x}(t)$ for $t=[k\tau  \quad (k+1)\tau]$, where $k\in \mathbb{N}$ and $\tau$ is the time horizon for which the trajectories are generated. This trajectory is considered under-collision if any state along its path is under collision. If the trajectory $\bm{x}(k\tau \leq t \leq (k+1)\tau )$ is collision-free, then it is accepted as a valid trajectory and the next look-ahead trajectory is generated starting from the state $x((k+1)\tau)$. This allows us avoid assuming hover as an initial condition for each look-ahead trajectory. The quadrotor executes the feasible trajectory $\bm{x}(t \leq k\tau)$ in parallel. If any state is outside the field of view, the trajectory generator keeps on checking it till the quadrotor gets closer to this state so that the camera can see it. Figure \ref{fig:traj_gen_cartoon} shows the trajectory generation method.

A look-ahead trajectory $\bm{x}(k\tau \leq t \leq (k+1)\tau)$ is generated either using the control objective $J_0(\bm{x},\bm{u})$ or $J_1(\bm{x},\bm{u})$. We use infinite horizon LQR controller to generate the trajectories for each objective. The control law is then applied for the time horizon $\tau$. 

Let a quadratic objective be, 
\begin{align*}
    \mathop{\textnormal{minimize}}_{\bm{u,x}} \: \: &J_i(\bm{u,x}) = \\ &\int_0^\infty \bm{x}(t)\bm{Q}_i\bm{x}(t) + \bm{u}(t)\bm{R}_i\bm{u}(t), \: i\in\{0,1\} \\
    \textnormal{subject to} \: &\dot{\bm{x}}(t) = \bm{A}\bm{x}(t) + \bm{B}\bm{u}(t), \quad \forall \: t \in [0\quad t_f] \\
                             &\bm{x}(0)=\bm{x}(k\tau), \quad \bm{x}(\infty) = \bm{x}_{ref}.
\end{align*}

The optimal input $\bm{u}(t)$ to the system is obtained by using Pontryagin's minimum principle and Hamilton-Jacobi-Bellman (H-J-B) equation. Here,
\begin{align}
    \bm{u}_i(t) = -\bm{R_i}^{-1}\bm{B}^T\bm{S}_i\bm{x}(t).
\end{align}
Here $\bm{S}_i$ is the solution to the steady state algebraic Riccati equation (ARE) given below,
\begin{align}
\label{eq:riccati_equation}
    \bm{S}_i\bm{A} + \bm{A}^T\bm{S}_i + \bm{Q_i} - \bm{S}_i\bm{B}\bm{R_i}^{-1}\bm{B}^T\bm{S}_i = 0.
\end{align}

Equation (\ref{eq:riccati_equation}) is solved for all $i$ prior to flight to save onboard computations. Two discrete modes are elaborated as follows.

\emph{Mode $l_0$ (Go-To-Goal):}
\par This mode is responsible for moving the quadrotor forward to its destination in an obstacle-free environment. The reference state is defined as $\bm{x}_{ref} \in \mathcal{X}_{goal}$. The objective function $\bm{J}_{0}$ is given by the matrices $\bm{Q}_0 = \textnormal{diag} (1, 0.1, 1, 0.1, 1, 0.1)$ and $\bm{R}_0 = \textnormal{diag} (3,3,3)$. These values are chosen such that the quadrotor can save more energy if there are no obstacles hindering it way. If any state in the look-ahead trajectory $\bm{x}(k\tau \leq t \leq (k+1)\tau)$ is under-collision, then it is discarded and is re-generated in the \textit{Escape} mode towards the nearest escape. The guard condition is hence given by $G(l_0, l_1) = \{\boldsymbol{x} (t = k\tau) : \exists \: \bm{x}(t) \in \mathcal{X}_{obs} \textnormal{ for any } t \in [k\tau \quad (k+1)\tau]\}$. 
 
\emph{Mode $l_1$ (Escape):}
\par Trajectories generated using this mode are designed to take the quadrotor toward the closest possible escape through an aggressive maneuver in order to have a clearer view of what is behind the obstacle. Trajectories generated using \textit{Go-To-Goal} mode tend to be curvy, and less agile. This fact poses problems while deviating the quadrotor from its path. Therefore, lesser weights are put on the input term in the objective function. The control law for the \textit{Escape} mode is computed using $\bm{Q}_1 = \textnormal{diag} (1, 0.1, 1, 0.1, 1, 0.1)$ and  $\bm{R}_1 = \textnormal{diag} (0.1,0.1,0.1)$. The reference state for this mode is $\bm{x}_{ref} = \bm{x}_{esc}$. The mode switches back to the \textit{Go-To-Goal} once the escape point is reached and the quadrotor may now see behind the obstacle. The guard associated with this transition is given by $G(l_1, l_0) = \{ \boldsymbol{x}(t = k\tau) : \boldsymbol{x} (t) = \boldsymbol{x}_{esc}  \}$.

\emph{Discussion:}
\par Problem (2) can be solved using the search or sampling-based path planning techniques \cite{liu2018search}. These techniques are particularly useful in the presence of non-convex constraints, $\bm{x}(t) \in \mathcal{X}_{free}$. However, these techniques are proven to be inefficient for planning in high dimensional spaces, because they have to rely on the expansion of large number of nodes. Also, the number of collision checks increases with the number of nodes. For the real-time planning, we are interested in techniques that require minimal collision checks to generate feasible paths. Our method generates dynamically feasible trajectories without requiring the planner to search the whole 3D workspace. However, an LQR problem is inherently unconstrained and the structure of the workspace is unknown. Therefore, we exploit the depth image space to perform the planning in $\mathcal{X}_{free}$. Regardless, the collision detector presented in Section \ref{sec:_perception_strategy}, can be used with several planning algorithms that require fast collision checks.

\section{IMPLEMENTATION}
\label{sec:implementation}

\subsection{Software Architecture}
 \label{sec:_software_hardware_architecture}
 
 \begin{figure}[!h]
\centering
\includegraphics[width=0.48\textwidth]{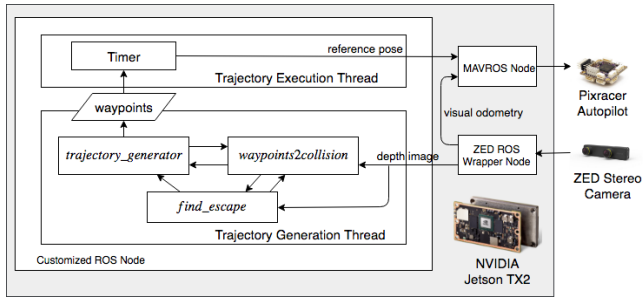}
\caption{\label{fig:architecture} Simplified software architecture.}
\end{figure}

\begin{figure*}[!ht]
  \begin{subfigure}[b]{0.32\textwidth}
    \includegraphics[width=\textwidth]{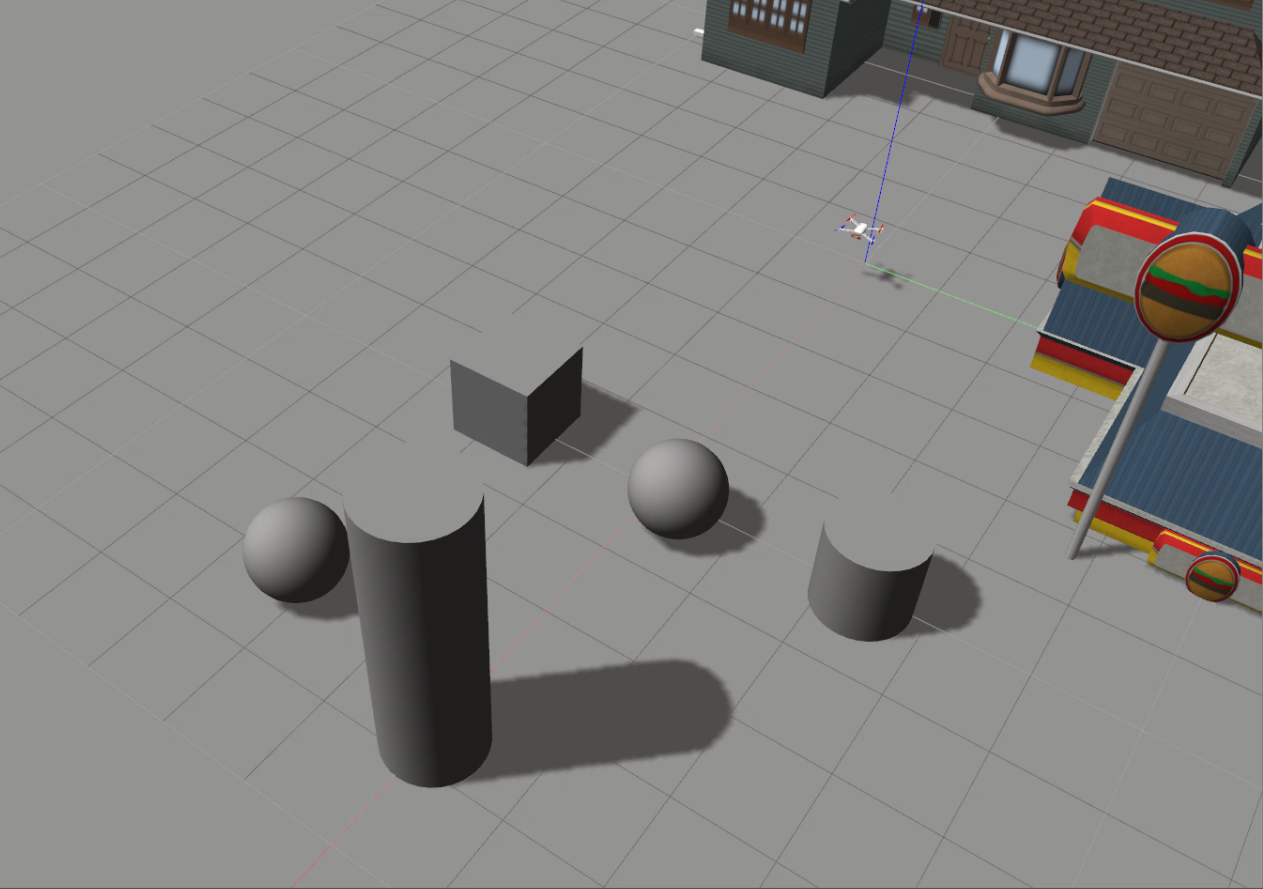}
    \caption{}
  \end{subfigure}
  \begin{subfigure}[b]{0.32\textwidth}
    \includegraphics[width=\textwidth]{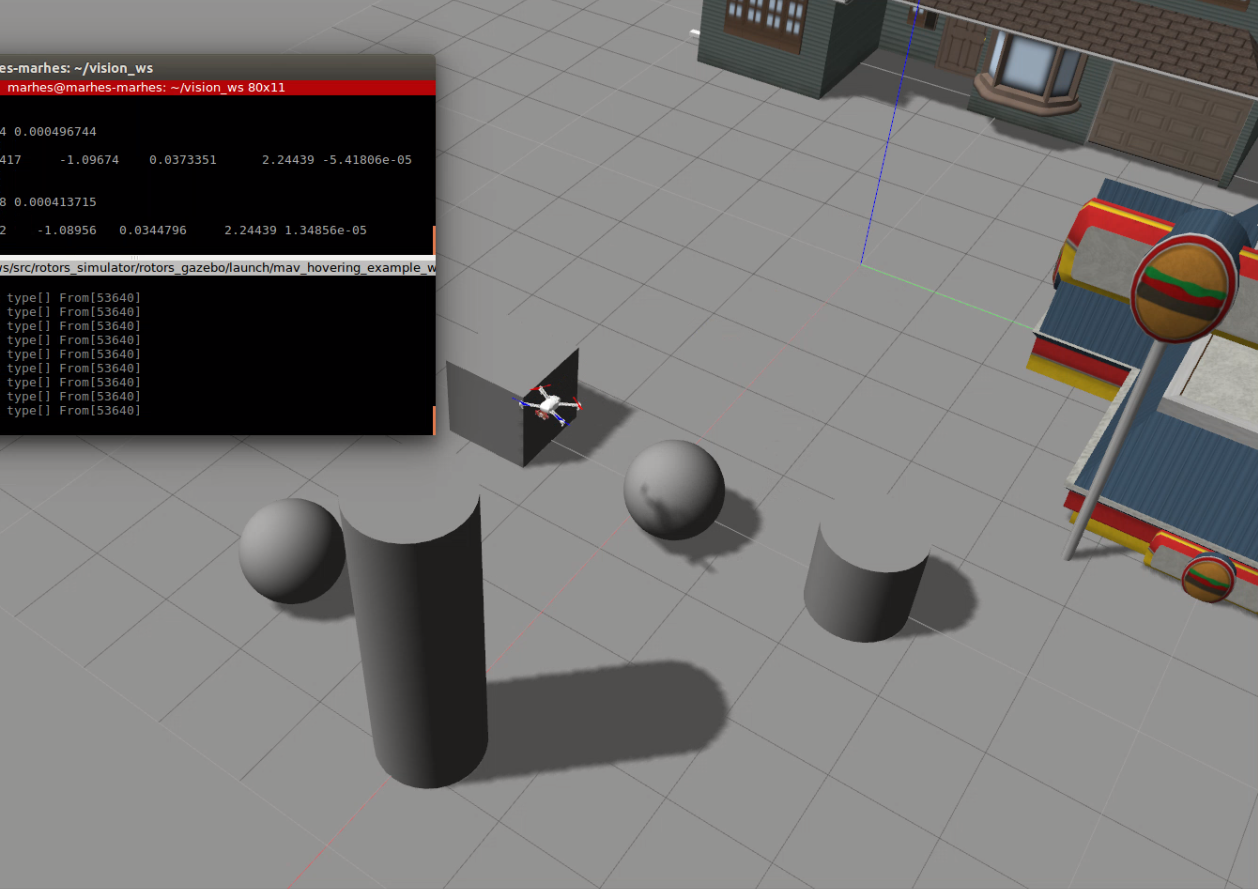}
    \caption{}
  \end{subfigure}
    \begin{subfigure}[b]{0.32\textwidth}
    \includegraphics[width=\textwidth]{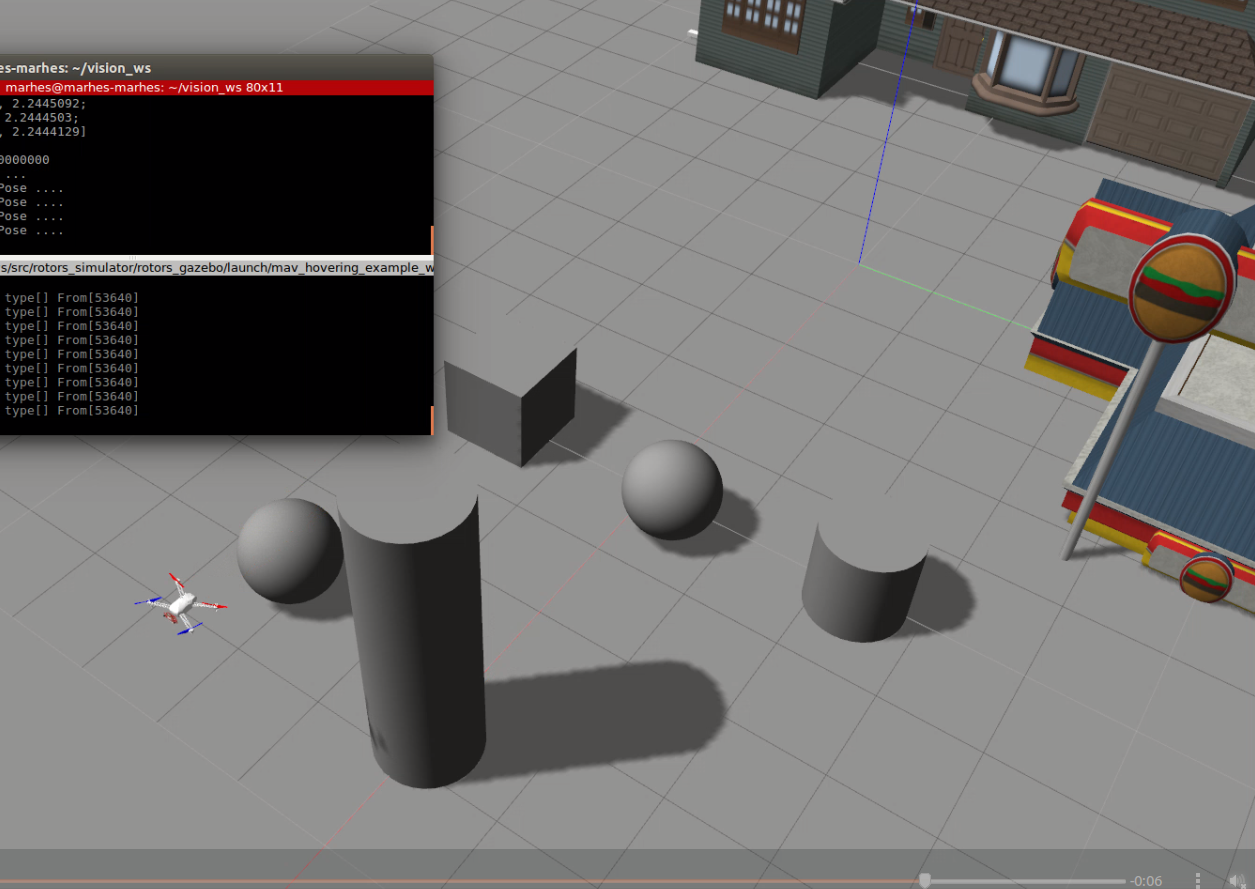}
    \caption{}
  \end{subfigure}
  \caption{\label{fig:gazebo_environment} Gazebo simulation environment. (a) and (c) show the quadrotor start and goal states for the maneuver respectively.}
\end{figure*}

\begin{figure} [!ht]
\centering
\includegraphics[width=0.5\textwidth]{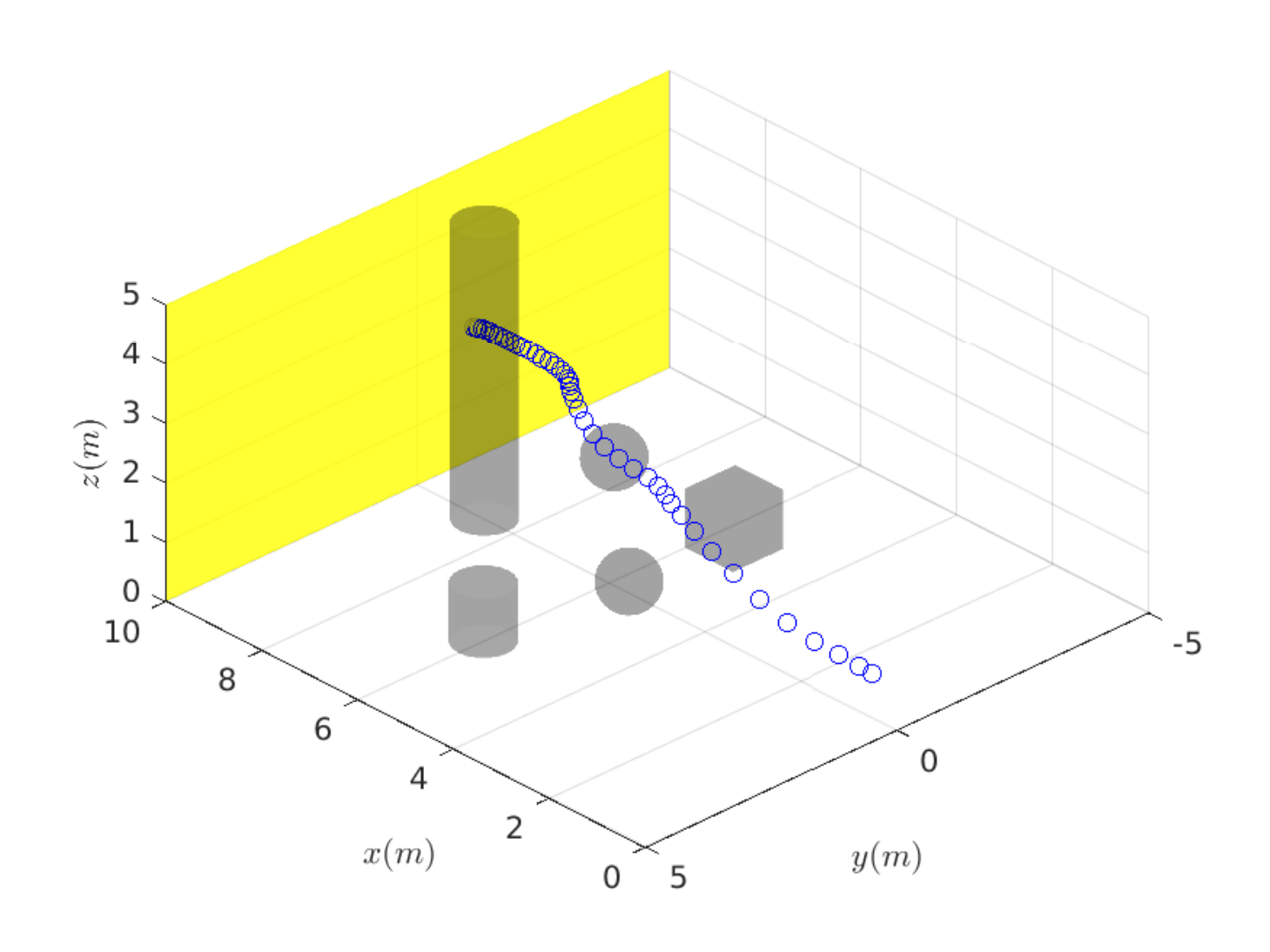}
\caption{\label{fig:final_traj_simulation} Feasible trajectory for Gazebo simulation setup. The region in yellow represents the goal.}
\end{figure}

 Multi-threading in Robot Operating System (ROS) is used with nodes and functions written in C++. Three primary functions in the trajectory generation thread include $find\_escape$, $waypoints2collision$, and $trajectory\_generator$. Let the initial quadrotor state be $\bm{x}_0$. The $trajectory\_generator$ function forms a trajectory $\bm{x}(t)$ for $t=[0 \quad \tau]$ under the \emph{Go-To-Goal} mode towards the goal state $x_{goal} \in \mathcal{X}_{goal}$. The trajectory $\bm{x}(t)$ is sent to $waypoints2collision$ function to perform collision checks. This function samples the trajectory with the sampling time $t_s$ and checks each state for collision. If the trajectory $\bm{x}(t)$ is collision-free and inside the quadrotors field of view, then it is accepted as a valid trajectory and the execution thread starts sending commands to the quadrotor. The execution thread runs at the sampling time of $t_s$. While the quadrotor (and hence the attached camera) is following $\bm{x}(t)$, the $trajectory\_generator$ generates the next trajectory $\bm{x}(t)$ for $t=[\tau \quad 2\tau]$ under the mode $l_0$ towards $\bm{x}_{goal}$. If any state along the trajectory $\bm{x}(t)$ for $t=[\tau \quad 2\tau]$ is under collision, the trajectory is discarded for this time duration. The function $find\_escape$ gets the information about the under collision state and finds an escape state $\bm{x}_{esc}$ around it. The $trajectory\_generator$ now generates the trajectory $\bm{x}(t)$ for $t=[\tau \quad 2\tau]$ towards the state $\bm{x}_{esc}$ under the $Escape$ mode. The $trajectory\_generator$ then waits for the quadrotor to physically reach the escape state before generating more trajectories so that the quadrotor can see past the obstacle. Once the quadrotor reaches the escape state, the $trajectory\_generator$ generates a trajectory in \emph{Go-To-Goal} and the process continues until the generated trajectories reach the goal state. This type of technique allows the trajectory generation and execution to run in parallel. The execution thread keeps on sending the reference states from the feasible trajectory to the autopilot. Meanwhile, the generation thread continues on appending the collision-free look-ahead trajectories to the already existing feasible trajectory. This type of multi-threading plays an important role in achieving the real-time maneuver. The simplified software architecture is shown in the Figure \ref{fig:architecture}.

\subsection{Simulations}

\begin{figure}[!ht]
\hspace*{-0.61cm}
\includegraphics[width=1.1\linewidth]{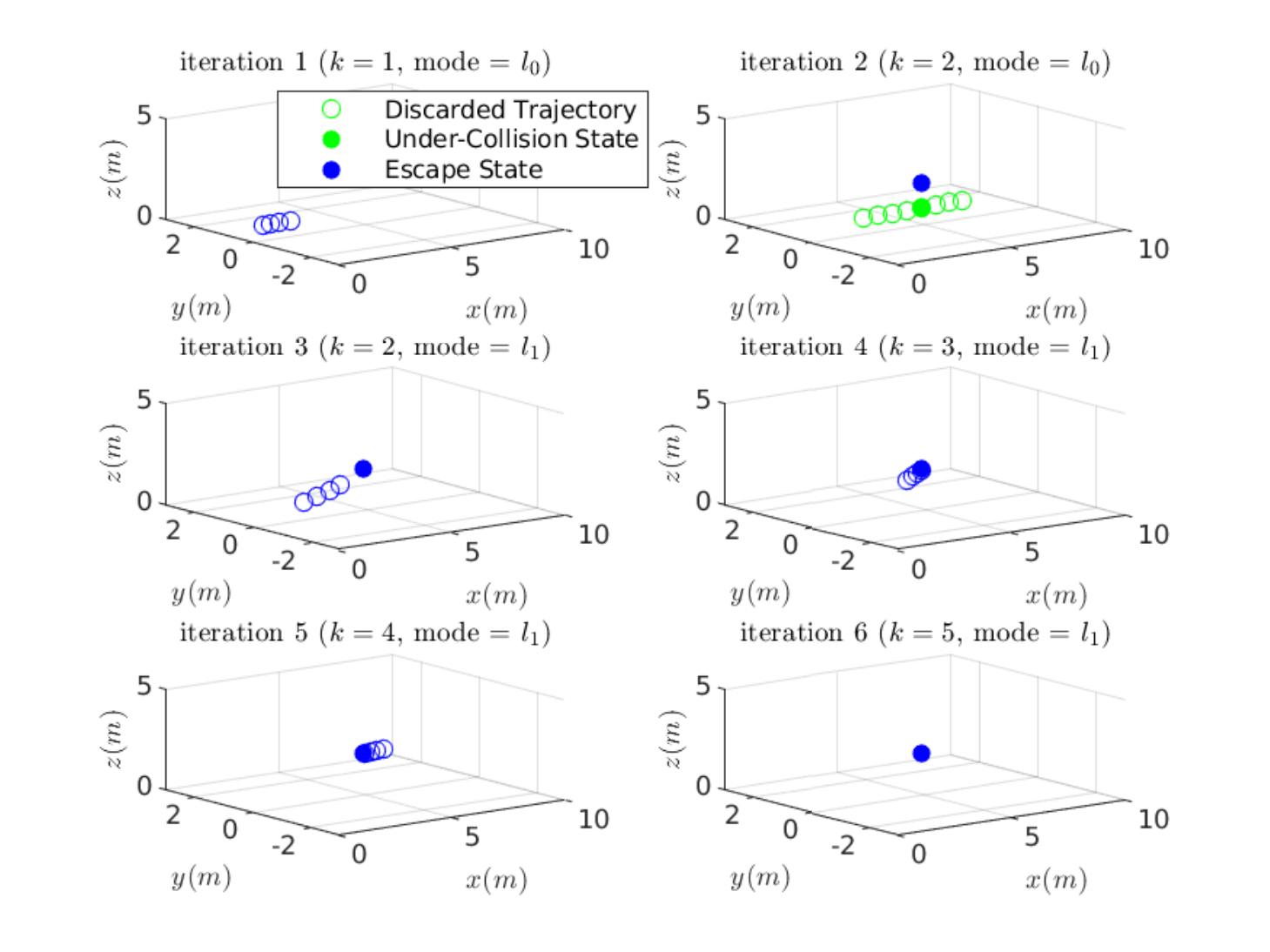}
\caption{\label{fig:traj_gen} First few iterations of look-ahead trajectories ($\bm{x}(k\tau \leq t \leq (k+1)\tau)$) generation for Gazebo simulation setup. The first look-ahead trajectory is generated in the iteration 1 under the mode $l_0$. Since this trajectory is collision-free, it is accepted as the feasible trajectory. The look-ahead trajectory generated in the iteration 2 using mode $l_0$ is under collision. This trajectory is discarded and a trajectory in mode $l_1$ is generated towards an escape in iteration 3. This trajectory is appended after the existing feasible trajectory. In the next few iterations, the trajectories are generated towards the escape.}
\end{figure}

We use Gazebo simulator with Robot Operating System (ROS) to simulate our framework. RotorS package is used \cite{rotors:2016} to simulate a quadrotor model with a forward facing depth camera. We tested several different obstacles configurations. One simulator setup is shown in Figure \ref{fig:gazebo_environment}. The quadrotor starts from an initial state as shown in the Figure \ref{fig:gazebo_environment}(a). The goal region $\mathcal{X}_{goal}$ is defined by a $y-z$ plane in $\mathcal{F_W}$. The quadrotor is expected to escape the cluttered environment and move forward towards the goal region. In this setup and the gains given in Section \ref{sec:_trajectory_generation}, the quadrotor took 8 s to reach its goal plane which was 10 m away from the quadrotor initial position. The step-by-step trajectory generation for a first few iterations is shown in the Figure \ref{fig:traj_gen}. The figure shows the generation of the look-ahead trajectories which are appended together to generate a feasible trajectory. The complete feasible trajectory after appending and discarding the look-ahead trajectories is shown in the Figure \ref{fig:final_traj_simulation}. 

\subsection{Hardware Experiments}

\begin{figure} [!h]
\centering
\includegraphics[width=0.47\textwidth]{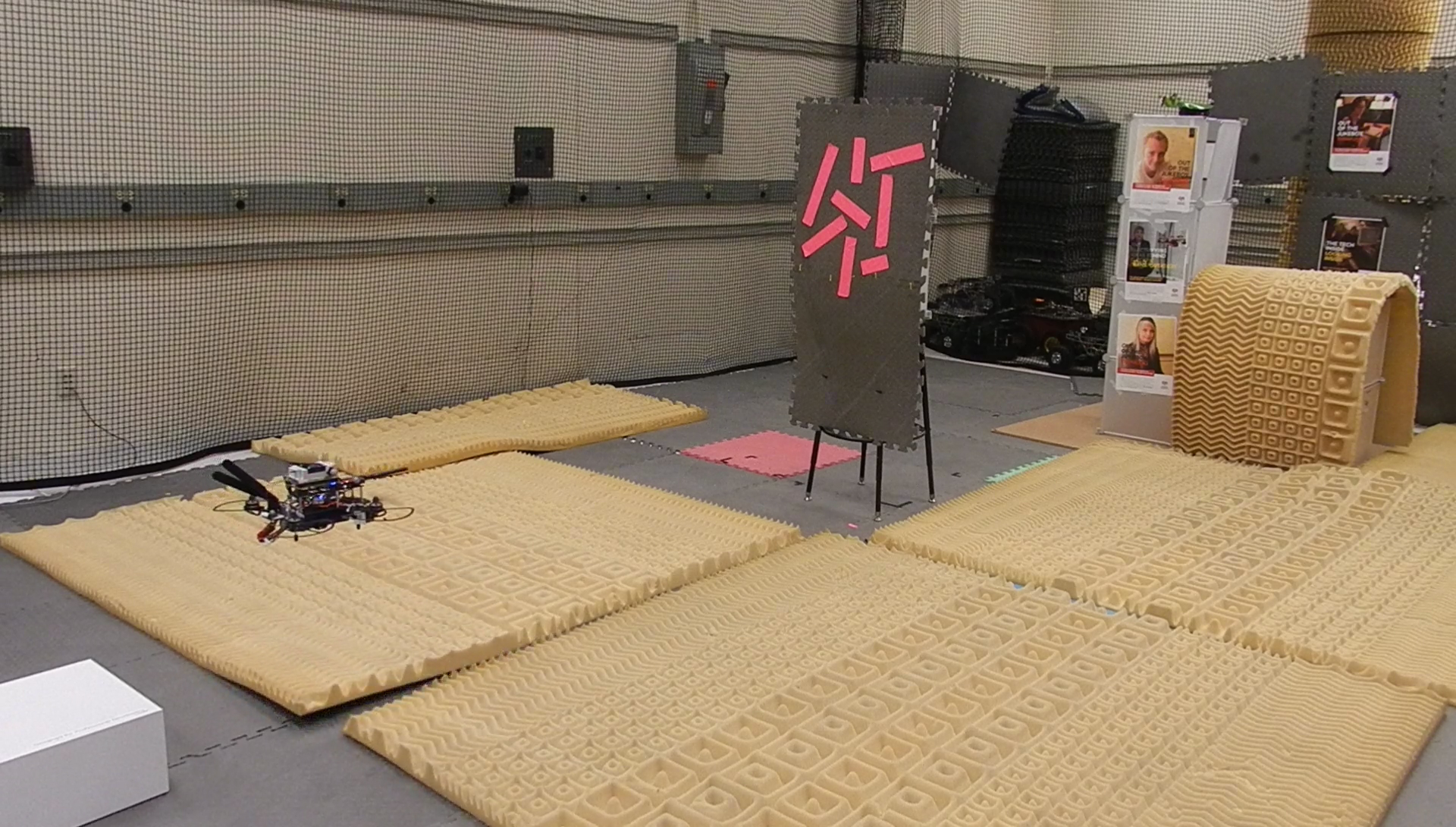}
\caption{\label{fig:hardware_setup1}Experimental setup 1.}
\end{figure}

\begin{figure} [!h]
\centering
\includegraphics[width=0.48\textwidth]{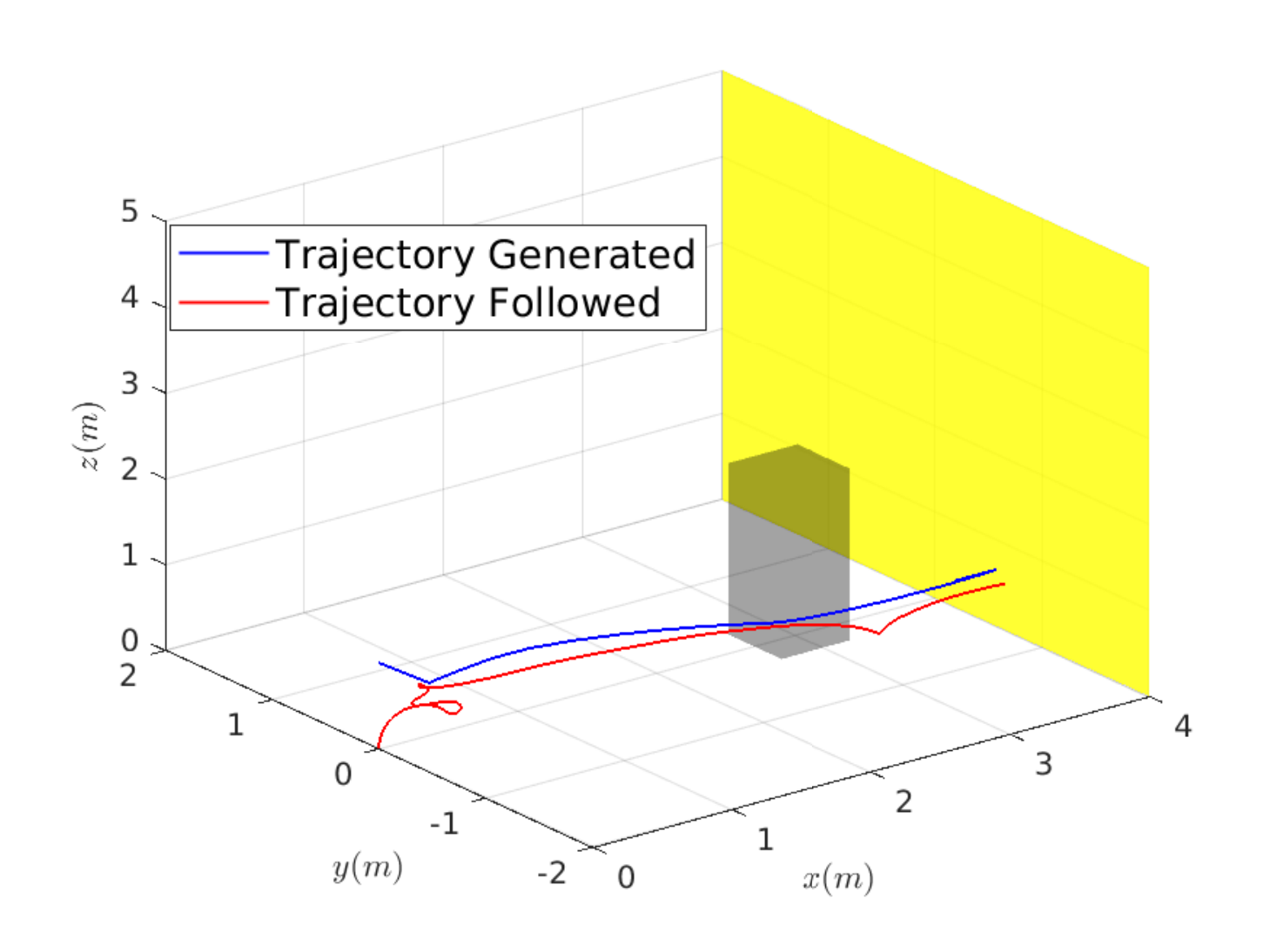}
\caption{\label{fig:final_traj_net_around} Trajectory results for experimental setup 1. The region in yellow represents the goal.}
\end{figure}

\begin{figure} [!h]
\centering
\includegraphics[width=0.47\textwidth]{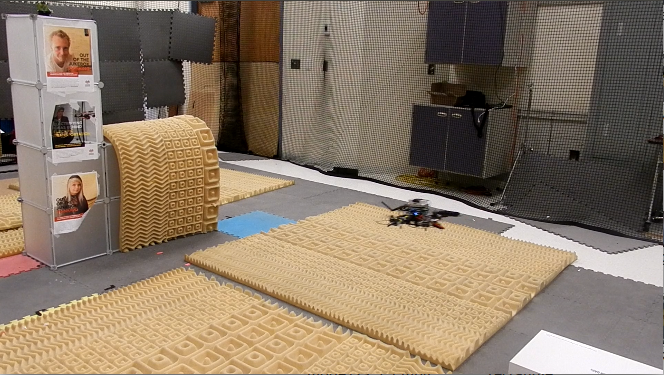}
\caption{\label{fig:hardware_setup2}Experimental setup 2.}
\end{figure}

\begin{figure} [!h]
\centering
\includegraphics[width=0.48\textwidth]{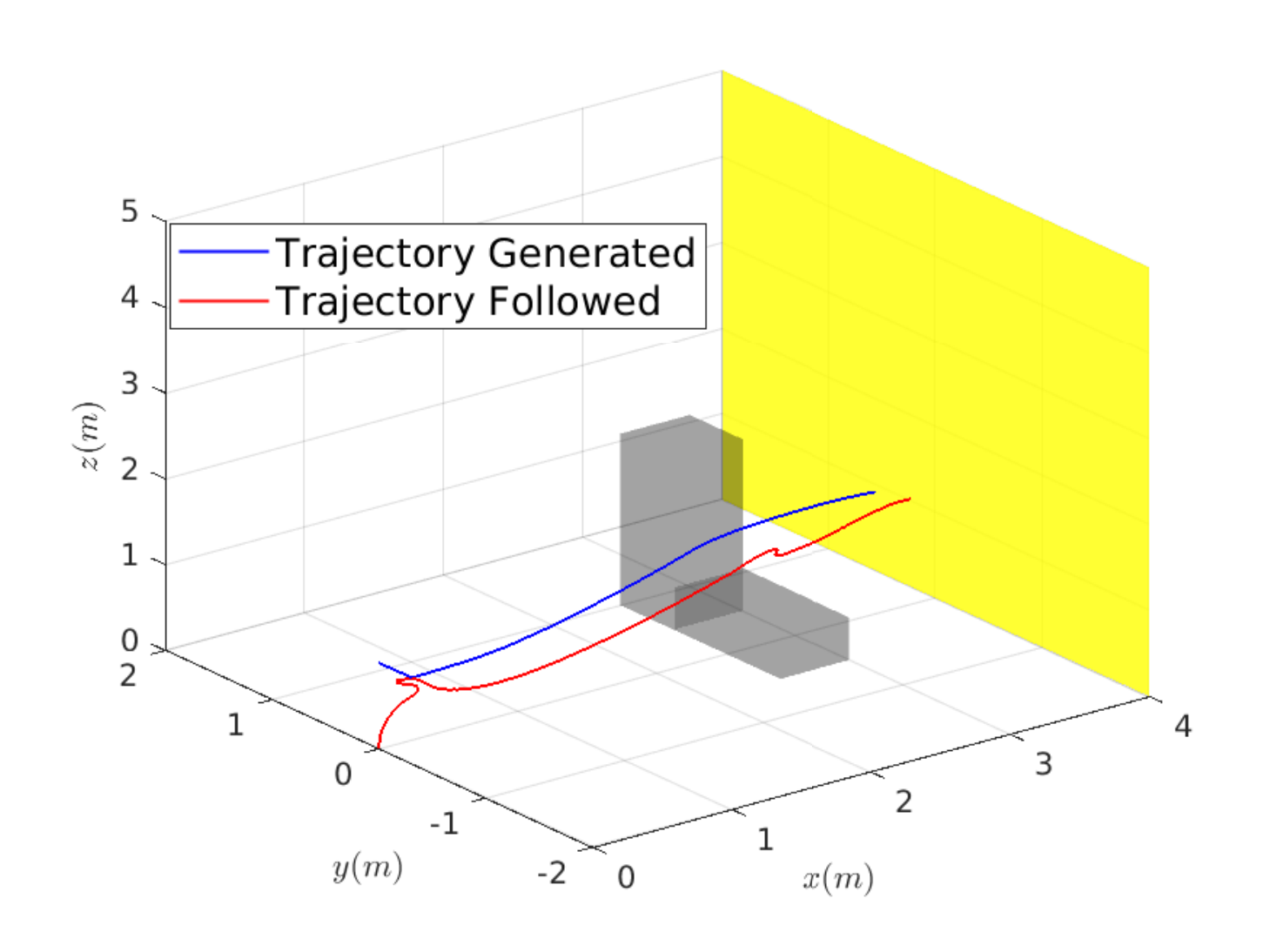}
\caption{\label{fig:final_traj_net_over} Trajectory results for experimental setup 2. The region in yellow represents the goal.}
\end{figure}

\begin{figure*}
  \begin{subfigure}[b]{0.32\textwidth}
    \includegraphics[width=\textwidth]{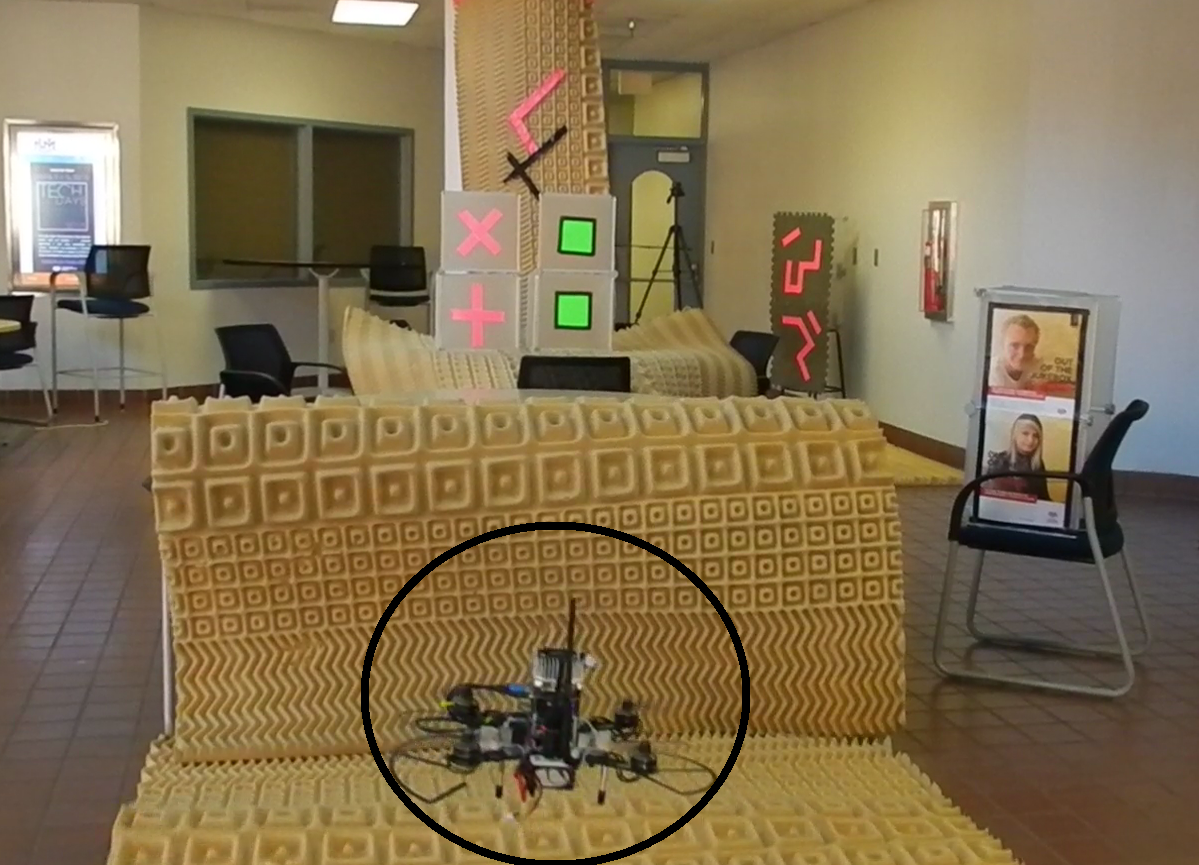}
    \caption{}
  \end{subfigure}
  \begin{subfigure}[b]{0.32\textwidth}
    \includegraphics[width=\textwidth]{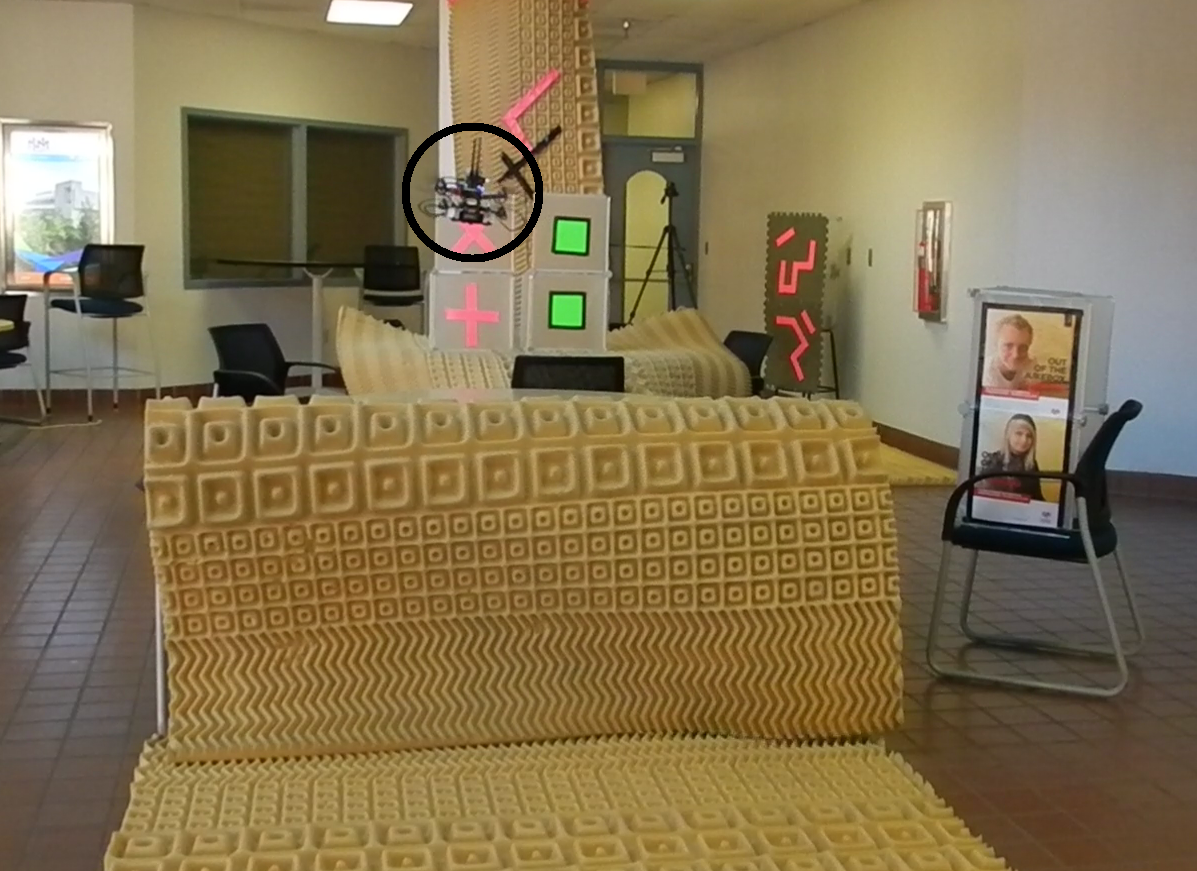}
    \caption{}
  \end{subfigure}
    \begin{subfigure}[b]{0.32\textwidth}
    \includegraphics[width=\textwidth]{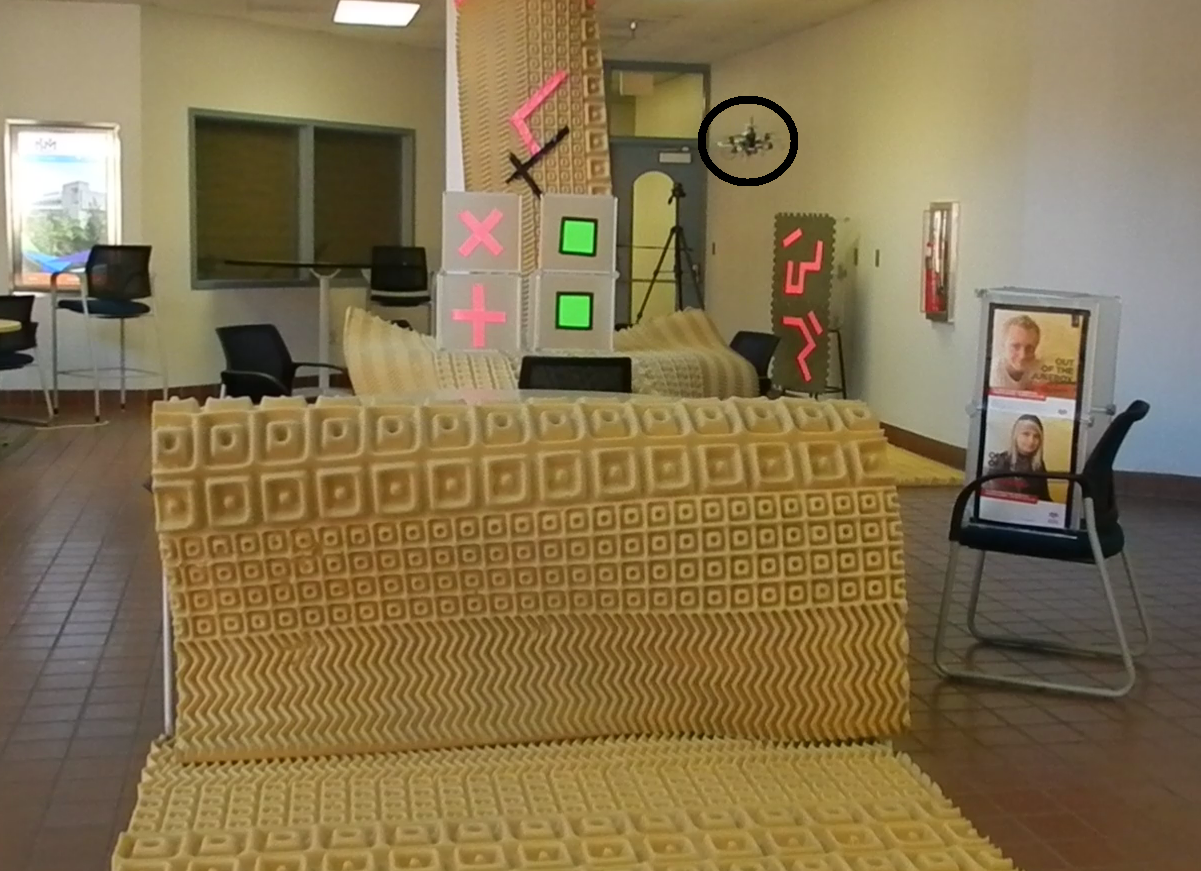}
    \caption{}
  \end{subfigure}
  \caption{\label{fig:hardware_setup3} Experimental setup 3. (a) and (c) show the quadrotor start and goal states for the maneuver respectively.}
\end{figure*}

\begin{figure} [!ht]
\centering
\includegraphics[width=0.5\textwidth]{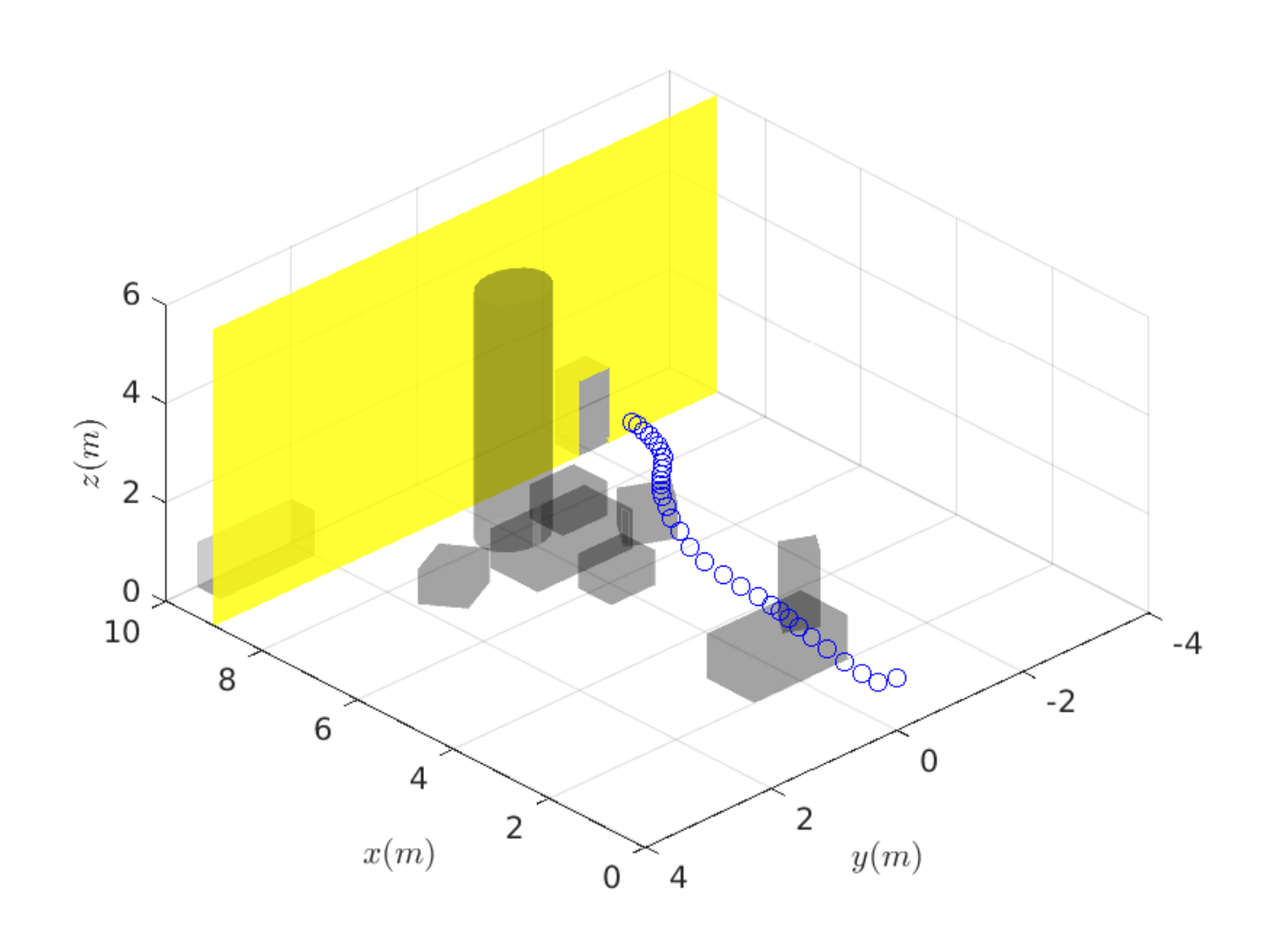}
\caption{\label{fig:final_traj_lobby} Feasible trajectory for experimental setup 3. The region in yellow represents the goal.}
\end{figure}

We verified our real-time planner on a real quadrotor (Lobo Drone). We equipped the quadrotor with a ZED mini stereo camera \cite{zedcamera} for obtaining the depth image stream and visual odometry. The state estimation is accomplished through an Extended Kalman Filter by fusing the visual odometry with the IMU data. All the processing and computations are performed onboard the Lobo Drone without any help of offboard sensors such as a motion capture system. All the code runs as ROS nodes on NVIDIA Jetson TX2 computer \cite{jetsontx2} onboard a Lumenier QAV250 quadrotor airframe with a PX4 based pixracer autopilot. This setup enables the system to execute autonomous maneuvers independently using only a forward facing camera. The experiments are performed in three different setups. In each setup, the quadrotor does not have any information about the workspace prior to flight and is only provided with the goal region. The Jetson TX2 computer fetch the 640x480 depth images at the rate of 30 FPS. The ZED mini stereo camera has the depth range of 10 m. The look-ahead trajectory is sampled at every $t_s$ = 0.2 s to perform the collision checks. The time horizon $\tau$ is kept at 0.8 s.

The first setup is shown in Figure \ref{fig:hardware_setup1}. The goal region is set at 4 m distance from the initial state. The quadrotor predicted a collision in a look-ahead trajectory by a thin obstacle. According to our escape strategy, the nearest possible escape was found towards the right of the obstacle. The quadrotor planned the look-ahead trajectories towards the escape and then towards the goal region. It took the quadrotor approximately 6 s to complete the maneuver. Figure \ref{fig:final_traj_net_around} shows the feasible trajectory generated and tracked by the planner and by the quadrotor respectively.

The second experiment is also performed in the same area but with two different obstacles placed together. The goal region is kept the same. The setup is shown in the Figure \ref{fig:hardware_setup2}. The quadrotor found an escape above the obstacle of lower height and planned the look-ahead trajectories to escape. The trajectory results for this experimental setup is shown in the Figure \ref{fig:final_traj_net_over}. The quadrotor took approximately 5 s to reach the goal state. It can also be noted that in using our method, the global information about the obstacles is not required to plan a feasible trajectory. Rather, only the obstacles causing a potential predicted collision matter. 

The third experiment is performed in a larger indoor lobby area in a more cluttered workspace. The quadrotor is left at at a starting hover state to escape the cluttered environment. Regardless of the obstacle geometries, the quadrotor was able to safely maneuver towards the goal plane that was 9 m away from the initial quadrotor state. The whole manuever took approximately 7 s to finish. The setup with start and goal states is shown in Figure \ref{fig:hardware_setup3}. The feasible trajectory for the flight is shown in Figure \ref{fig:final_traj_lobby}.

One challenge in using depth image based perception is that the typical stereo cameras are sometimes not able to compute the depth at each image pixel. The ZED stereo camera, however, provides a highly filtered depth image stream over a ROS interface. The problem sometimes occurs when the camera tries to compute the depth of a very intense light source like a light bulb. In such rare circumstances the corresponding pixel depths are declared as the maximum allowed depth range of the camera. If the light source is at a sufficient distance from the camera, its projection and hence the noise in the whole image gets negligible. 

\section{CONCLUSION}
\label{sec:conclusion}
This paper has proposed a technique for accomplishing fast autonomous quadrotor navigation through unstructured cluttered environments using only a forward facing stereo camera as a primary perception source. We exploited the depth image space to perform fast collision checks and to generate collision-free dynamically feasible trajectories in real-time. Also, we proposed the software and hardware architectures to effectively implement the resulting vision-based planner onboard a quadrotor UAV. Finally, we verified the approach in simulation and hardware experiments.
 \par

The Videos for simulations and experiments can be found at \url{https://youtu.be/zO-yT8K3SCg}. The code is also open source and can be found at \url{https://github.com/shakeebbb/vision_pkg}.

\bibliographystyle{IEEEtran}

\bibliography{./bibliography/bibliography}

\end{document}